\documentclass[letterpaper]{article} 
\usepackage{aaai24}  
\usepackage[title]{appendix}
\usepackage{amssymb}
\usepackage{amsmath}
\usepackage{booktabs}
\usepackage{algorithm,algpseudocode}
\usepackage{multirow}
\usepackage{makecell}
\usepackage{tikz}
 \usepackage{enumitem}
 \usepackage{cite}
 \usepackage{dsfont}

\usepackage{color}
\usepackage{xcolor}
\usepackage{soul}
\definecolor{carnelian}{rgb}{0.7, 0.11, 0.11}
\newcommand{\ifcomments}{\iftrue}

\newcommand{\zy}[1]{\textcolor{red}}

\definecolor{Gray}{gray}{0.9}
\definecolor{LightCyan}{rgb}{0.88,1,1}

\newcolumntype{a}{>{\linecolor{Gray}}c}
\newcolumntype{b}{>{\linecolor{red!20}}c}

\usepackage{color}
\usepackage{xcolor}
\usepackage{aaai24}  
\usepackage{times}  
\usepackage{helvet}  
\usepackage{courier}  
\usepackage[hyphens]{url}  
\usepackage{graphicx} 
\usepackage{natbib}  
\usepackage{caption} 
\usepackage{algorithm}
\usepackage{algorithm,algpseudocode}
\newcommand{\name}{{\textsc{VQAttack}}}

\newcolumntype{a}{>{\columncolor{blue!10}}c}
\usepackage{newfloat}
\usepackage{listings}
\DeclareCaptionStyle{ruled}{labelfont=normalfont,labelsep=colon,strut=off} 
\floatstyle{ruled}
\newfloat{listing}{tb}{lst}{}
\floatname{listing}{Listing}
\title{VQAttack: Transferable Adversarial Attacks on Visual Question Answering via Pre-trained Models}

\author{
    Ziyi Yin\textsuperscript{\rm 1},\;
    Muchao Ye\textsuperscript{\rm 1},\;
    Tianrong Zhang\textsuperscript{\rm 1},\;
    Jiaqi Wang\textsuperscript{\rm 1},\;
    Han Liu\textsuperscript{\rm 2}\\
    Jinghui Chen\textsuperscript{\rm 1},\;
    Ting Wang\textsuperscript{\rm 3},\;
    Fenglong Ma\textsuperscript{\rm 1}\thanks{Corresponding author.}
}
\affiliations {
    \textsuperscript{\rm 1}The Pennsylvania State University\\
    \textsuperscript{\rm 2}Dalian University of Technology\\
    \textsuperscript{\rm 3}Stony Brook University \\
    \{ziyiyin, muchao, tbz5156, jcz5917, fenglong\}@psu.edu\\ 
    liu.han.dut@gmail.com,\;
    twang@cs.stonybrook.edu
}

\begin{document}
\maketitle

\begin{abstract}
Visual Question Answering (VQA) is a fundamental task in computer vision and natural language process fields. Although the ``pre-training \& finetuning'' learning paradigm significantly improves the VQA performance, the adversarial robustness of such a learning paradigm has not been explored. 
In this paper, we delve into a new problem: using a pre-trained multimodal source model to create adversarial image-text pairs and then transferring them to attack the target VQA models.
Correspondingly, we propose a novel {\name} model, which can iteratively generate both image and text perturbations with the designed modules: the large language model (LLM)-enhanced image attack and the cross-modal joint attack module.
At each iteration, the LLM-enhanced image attack module first optimizes the latent representation-based loss to generate feature-level image perturbations. Then it incorporates an LLM to further enhance the image perturbations by optimizing the designed masked answer anti-recovery loss. 
The cross-modal joint attack module will be triggered at a specific iteration, which updates the image and text perturbations sequentially. Notably, the text perturbation updates are based on both the learned gradients in the word embedding space and word synonym-based substitution.
Experimental results on two VQA datasets with five validated models demonstrate the effectiveness of the proposed {\name} in the transferable attack setting, compared with state-of-the-art baselines. This work reveals a significant blind spot in the ``pre-training \& fine-tuning'' paradigm on VQA tasks. Source codes will be released.
\end{abstract}
\section{Introduction}

Visual Question Answering (VQA) is dedicated to extracting essential information from images to formulate responses to textual queries. While this application has proven to be highly versatile across various domains, including recommendation systems~\cite{recommend}, medicine~\cite{medvqa}, and robotics~\cite{robotics}, the exploration of VQA system robustness remains a challenging endeavor. Current research primarily revolves around investigating the robustness of \emph{end-to-end trained VQA models} through the development of effective attack methodologies, exemplified by Fool-VQA~\cite{foolvqa} and TrojVQA~\cite{trojvqa}. However, models trained end-to-end often exhibit inferior performance compared to the prevalent ``pre-training \& fine-tuning'' paradigm.
\begin{figure}[t]
         \centering
         \includegraphics[width=0.47\textwidth]{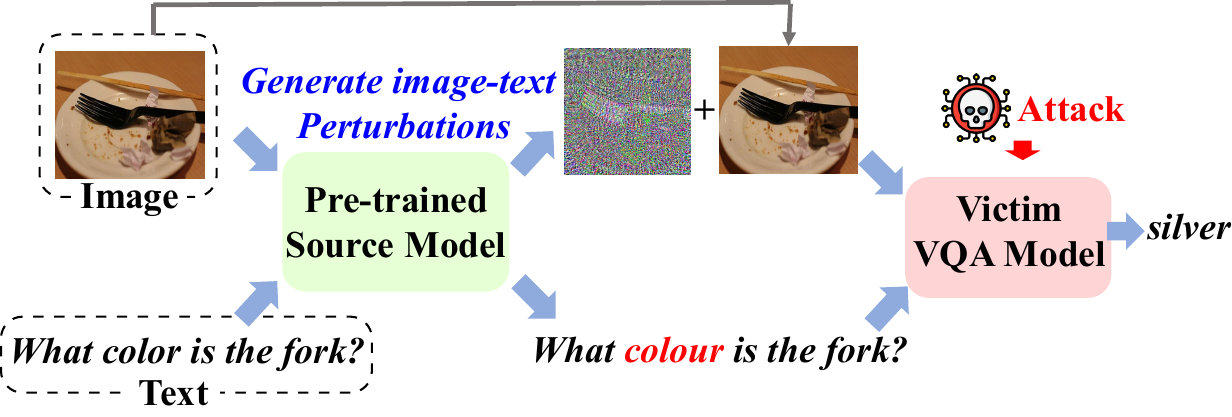}
         \caption{An example of Transferable adversarial attacks on VQA via pre-trained models.}
         \label{fig:intro}
\end{figure}
\begin{figure*}[t]
         \centering
         \includegraphics[width=0.95\linewidth]{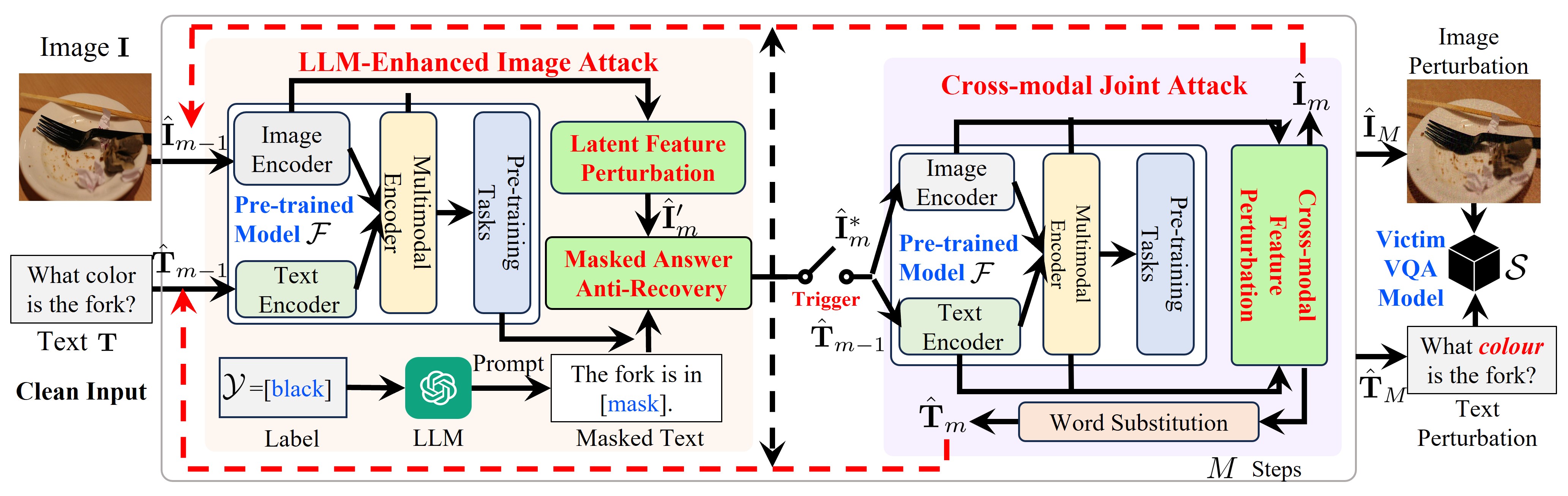}
         \caption{Overview of the proposed {\name}.}
         \label{fig:model}
\end{figure*}
Within this paradigm, models are initially pre-trained on extensive collections of image-text pairs from the public domain, facilitating the acquisition of intermodal relationships. Subsequently, the models undergo fine-tuning using specific VQA datasets to enhance their performance on downstream tasks. This instructional framework has yielded commendable predictive accuracy~\cite{vilt, vlmo, albef}. Nevertheless, the aspect of \textbf{adversarial robustness} within the context of the VQA task, as governed by this paradigm, remains insufficiently explored.

This attack scenario presents notable complexities, which arise from the following two fundamental aspects:
\begin{itemize}
    \item \textbf{C1 -- Transferability across models}. The challenge here involves the transferability of adversarial attacks across distinct models. An example is shown in Figure~\ref{fig:intro}. Pre-trained source models and victim target VQA models are usually trained for dissimilar tasks and trained on separate datasets. Furthermore, their structural disparities may result from variations introduced during fine-tuning. While the concept of transferability has been widely validated in the context of image models~\cite{pgd, diattack}, such property within the domain of pre-trained models has yet to be comprehensively explored.
    \item \textbf{C2 -- Joint attacks across different modalities}. Our task is centered around a multi-modal problem, necessitating the introduction of perturbations to both images and textual questions to achieve improved performance. Although previous methodologies have effectively devised attack strategies for each individual modality~\cite{pgd, bertattack}, the intricate challenge lies in the simultaneous optimization of perturbations on images with continuous values and textual content characterized by discrete tokens. This joint attack task continues to pose a significant hurdle that requires innovative solutions.
\end{itemize}

To address these challenges, we propose a novel method named {\name} to explore the adversarial transferability between pre-trained source and victim target VQA models.
As shown in Figure~\ref{fig:model}, the proposed {\name} generates image and text perturbations solely based on the pre-trained source model $\mathcal{F}$ with a novel multi-step attacking framework. 
After initializing the input image-text pair $(\mathbf{I}, \mathbf{T})$, {\name} will iteratively generate both image and text perturbations at each iteration $m$ via two key modules: large language model (LLM)-enhanced image attack and cross-modal joint attack.

In the \textbf{LLM-enhanced image attack} module, {\name} first follows existing work~\cite{coattack,SSP} to minimize the similarity of latent features between the clean and perturbed input and then uses the clipping technique to obtain the image perturbation $\hat{\mathbf{I}}_m^\prime$. To further enhance the transferability of attacks, {\name} introduces a new masked answer anti-recovery loss with the help of ChatGPT~\cite{gpt4}, which differs from the existing latent feature-level attack by involving the correct answer label $\mathcal{Y}$ during the perturbation generation. The LLM-enhanced image attack module will be executed at each iteration, and the output from this module is denoted as $\hat{\mathbf{I}}_m^*$.

Due to the discrete nature of text data and the limited number of informative words in each text input, attacking the text at every iteration might not be necessary or beneficial for perturbation generation. Consequently, the \textbf{cross-modal joint attack} module will be triggered when $m$ satisfies a specific condition. During this stage, {\name} first updates the perturbation of the image (i.e., $\hat{\mathbf{I}}_m^{*}$) via the cross-modal feature perturbation and the clipping technique. It then updates the text perturbation $\hat{\mathbf{T}}_m$ using the learned gradients and word synonym-based substitution in the word embedding space. 

{\name} will return the output after iterating for $M$ steps as the final adversarial image-text pair, i.e., $(\hat{\mathbf{I}}_{M}, \hat{\mathbf{T}}_{M})$, which will be used to attack the victim VQA model $\mathcal{S}$. Our contributions can be summarized as follows:

\begin{itemize}
    \item  To the best of our knowledge, this is the first study on the adversarial robustness of the VQA task under the ``pre-training $\&$ fine-tuning'' paradigm. It does not only discuss the robustness of this paradigm but, more importantly, probes the potential security concern under a realistic scenario.
     \item We propose {\name}, which is a novel method to generate adversarial image-text pairs on the pre-trained vision language models. It consists of two novel modules, utilizes an LLM to generate masked text, and enables the iterative joint attack between image and text modalities.
    \item Five pre-trained models and two VQA datasets are involved in our experiment. Experimental results verify the effectiveness of the proposed {\name} under the transferable attack setting. 
\end{itemize}
\section{The Proposed {\name}}

\subsection{Problem Formulation}
We use $\mathcal{F}$ to denote the publicly available pre-trained VL source model and $\mathcal{S}$ to represent the victim VQA target model. 
The goal of the transferable VQA attack is to generate an adversarial image-text pair $(\hat{\mathbf{I}},\hat{\mathbf{T}})$ on the pre-trained source model $\mathcal{F}$ using the clean input $(\mathbf{I}, \mathbf{T})$, which will make the target victim model $\mathcal{S}$ have a wrong prediction, i.e., $S(\hat{\mathbf{I}},\hat{\mathbf{T}})\notin \mathcal{Y}$, where $\mathcal{Y}$ is the set of correct answers. 

However, the victim model $\mathcal{S}$ in our setting is a black-box, arbitrary, and unknown model, and the only model that we can access is the pre-trained source model $\mathcal{F}$. Let $\mathcal{G}$ denote the proposed transferable attack strategy {\name}. We use the following function to generate an adversarial image-text pair $(\hat{\mathbf{I}},\hat{\mathbf{T}})$:
\begin{equation}
    (\hat{\mathbf{I}},\hat{\mathbf{T}}) = \mathcal{G}(\mathcal{F}, ({\mathbf{I}},{\mathbf{T}}), M, \sigma_i, \sigma_s), 
\end{equation}
where $\mathcal{G}$ is an iterative attacking function, and $M$ is the number of iterations. $\sigma_i$ and $\sigma_s$ are two hyperparameters to control the quality of adversarial images and text, which are defined as follows:
\begin{equation}\label{eq:constraints}
    \begin{array}{l}
        \text{Image:}\quad \lVert \hat{\mathbf{I}}-\mathbf{I}\rVert_{\infty}<\sigma_i, \\
        \text{Text:}\quad Cos(U(\hat{\mathbf{T}}),U(\mathbf{T}))>\sigma_s.
    \end{array}
\end{equation}
For an adversarial image $\hat{\mathbf{I}}$, we add pixel-level perturbations under the $L_\infty$-norm distance. The distance threshold is set to $\sigma_i$. For an adversarial sentence $\hat{\mathbf{T}}$, we replace words with their synonyms and enforce a semantic similarity constraint $\sigma_s$, which is implemented through the cosine similarity $Cos(\cdot,\cdot)$ between the sentence embeddings $U(\hat{\mathbf{T}})$ and $U(\mathbf{T})$. Here, $U(\cdot)$ represents the universal sentence encoder~\cite{USE}, which has been widely adopted in text attack methods~\cite{textbugger,textfooler,CLARE}. 

\begin{algorithm}[t]
	\renewcommand{\algorithmicrequire}{\textbf{Input:}}
        
	\renewcommand{\algorithmicensure}{\textbf{Output:}}
	\caption{The proposed {\name}}
	\label{alg:overview}
        \begin{algorithmic}[1]
        \Require A pre-trained source model $\mathcal{F}$, a clean image-text pair $(\mathbf{I},\mathbf{T})$ and the ground-truth label $\mathcal{Y}$, step-size $\epsilon$, prompt $\mathcal{P}$, LLM;
        \Require  Perturbation budget $\sigma_i$ on image, $\sigma_s$ on text, and the number of total iterations $M$. 

        \State \textbf{Initialization} $\hat{\mathbf{I}}_0=\mathbf{I}+\delta$, $\delta\in\mathcal{U}(0,1)$, ;\ $\hat{\mathbf{T}}_0=\mathbf{T}$, and use BERT model to generate candidate token set $\mathcal{C}$.
        \For{$m = 1$ to $M$}
        
        \State {// \emph{LLM-enhanced Image Attack}}
        \State \textcolor{gray}{// Perturbation Generation with Latent features}
         \State Calculate $\nabla_{i}\mathcal{L}_f^m$ via Eq.~\eqref{eq:feature_attack} using $(\hat{\mathbf{I}}_{m-1},\hat{\mathbf{T}}_{m-1})$;
        \State $\hat{\mathbf{I}}_m^\prime={\rm clip}_{\sigma_i}(\hat{\mathbf{I}}_{m-1}+\epsilon{\rm sign}(\nabla_{i}\mathcal{L}_{T}))$;

        \State \textcolor{gray}{// LLM-based Perturbation Enhancement}
        \State Masked text generation with LLM using $\hat{\mathbf{T}}_{m-1}$, label $\mathcal{Y}$, and prompt $\mathcal{P}$; 
        \State Calculate gradiants $\nabla_{i}\mathcal{L}_{a}^m$ via Eq.~\eqref{eq:llm_loss};
        
        \State $\hat{\mathbf{I}}_m^*={\rm clip}_{\sigma_i}(\hat{\mathbf{I}}_m^\prime+\epsilon{\rm sign}(\nabla_{i}\mathcal{L}_{a}^m))$; 
       
        \State {// \emph{Cross-modal Joint Attack}}
        
        \If{$m \mod \lfloor\frac{M}{|\mathcal{W}|+1}\rfloor=0$}
        \State \textcolor{gray}{// Image Perturbation Update}
        \State Calculate $\nabla_{i}\mathcal{L}_{c}^m$ via Eq.~\eqref{eq:cross_attack} using $(\hat{\mathbf{I}}_m^*$ $\hat{\mathbf{T}}_{m-1})$;
        \State $\hat{\mathbf{I}}_m={\rm clip}_{\sigma_i}(\hat{\mathbf{I}}_m^*+\epsilon{\rm sign}(\nabla_{i}\mathcal{L}_{c}^m))$;

        \State \textcolor{gray}{// Text Perturbation Update}
        \State Latent word embedding estimation via Eq.~\eqref{eq:word_embedding};
        \State Obtain the synonym ranks $\mathcal{R}(\mathcal{C})$ according to Eq.~\eqref{eq:ranking};
        \State Conduct synonym substitution to obtain $\hat{\mathbf{T}}_m$;
        \Else { $\hat{\mathbf{I}}_m = \hat{\mathbf{I}}^*_m$, $\hat{\mathbf{T}}_m = \hat{\mathbf{T}}_{m-1}$;}
        \EndIf
        \EndFor
    \State \textbf{return} $(\hat{\mathbf{I}}_M,\hat{\mathbf{T}}_M)$
    
\end{algorithmic}
\end{algorithm}

\subsection{Overview}
As shown in Figure~\ref{fig:model}, the proposed {\name} $\mathcal{G}$ first initializes the input pair $(\mathbf{I}, \mathbf{T})$ as $(\hat{\mathbf{I}}_0, \hat{\mathbf{T}}_0)$, and then updates $(\hat{\mathbf{I}}_m, \hat{\mathbf{T}}_m)$ at each iteration $m$ through the proposed large language model (LLM)-enhanced image attack and cross-modal joint attack until the maximum iteration $M$. The final output $(\hat{\mathbf{I}}_M, \hat{\mathbf{T}}_M)$ is then used to attack the victim model $\mathcal{S}$. 
Algorithm~\ref{alg:overview} shows the algorithm flow of the proposed {\name}.
Next, we provide the details of our model design step by step.

\subsection{Initialization}
As shown in Algorithm~\ref{alg:overview} line 1, for the input image $\mathbf{I}$, we follow the Projected Gradient Decent
(PGD)~\cite{pgd} method to initialize $\hat{\mathbf{I}}$ by adding noise $\delta$ sampled from the Gaussian distribution $\mathcal{U}$, i.e., $\hat{\mathbf{I}}_0=\mathbf{I}+\delta$, where $\delta\in\mathcal{U}(0,1)$. For the text modality, we directly use the original input as the initialization, i.e., $\hat{\mathbf{T}}_0 = \mathbf{T}$.

Intuitively, the initialized pair $(\hat{\mathbf{I}}_0, \hat{\mathbf{T}}_0)$ can serve as the initial input for the cross-modal joint attack module, where iterative updates are performed on $(\hat{\mathbf{I}}_m, \hat{\mathbf{T}}_m)$ at each iteration. However, it is worth noting that this seemingly straightforward approach may not yield adversarial examples of high quality for effectively attacking the targeted model $\mathcal{S}$.

One aspect to consider is the intrinsic disparity between the numerical pixel representation of the input image $\mathbf{I}$ and the sequence-based nature of the input text $\mathbf{T}$. Frequent perturbations to the discrete $\mathbf{T}$ can often result in significant gradient fluctuations, which could subsequently adversely impact the perturbation of the numerical $\mathbf{I}$. As such, strictly coupling the updates of these two modalities throughout the entire attack process may not be the most optimal strategy.
Besides, the input text $\mathbf{T}$ is typically characterized by a relatively short average length\footnote{According to our investigation on the VQAv2 validation set, each sentence is only composed of an average of 6.21 words.}, containing only a limited number of informative words. This leads us to recognize that attacking the text at every iteration might not be necessary or beneficial.

It is due to these considerations that we put forth a novel module, namely the LLM-enhanced image attack. This module is designed to first learn an effective image perturbation independently, subsequently followed by a collaborative update of both image and text perturbations iteratively.

\subsection{LLM-enhanced Image Attack}

\subsubsection{Perturbation Generation with Latent Features}
Several approaches have been proposed to generate the image perturbations using pre-trained models, such as Co-Attack~\cite{coattack} and BadEncoder~\cite{badencoder}. The goal of these approaches is to minimize the similarity between the latent features learned by the pre-trained model $\mathcal{F}$ using the clean $\mathbf{I}$ and the perturbed $\hat{\mathbf{I}}_{m-1}$ at each iteration $m$, respectively. 

Most multimodal VL pre-trained models such as ViLT~\cite{vilt} and VLMO~\cite{vlmo} usually consist of three encoders to learn latent features, including an image encoder, a text encoder, and a multimodal encoder. To generate the perturbation of $\hat{\mathbf{I}}_m$, we first follow existing work to update the image perturbation by minimizing the following loss function:
\begin{equation}\label{eq:feature_attack}
    \mathcal{L}_{f}^m = \underbrace{\sum_{i=1}^{L_p} \sum_{j=1}^{D_p}Cos(\mathbf{f}_{i,j}^{p},\;\hat{\mathbf{f}}_{i,j}^{p})}_{\text{image encoder}} + \underbrace{\sum_{i=1}^{L_q}\sum_{j=1}^{D_q}Cos(\mathbf{f}_{i,j}^{q},\;\hat{\mathbf{f}}_{i,j}^{q})}_{\text{multimodal encoder}},
\end{equation}
where $L_p$ and $L_q$ denote the number of layers in the image encoder and multimodal encoder, respectively. $D_p$ and $D_q$ represent the number of input tokens of the image encoder and multimodal encoder. For the image encoder, the input tokens are image patches; and the multimodal encoder takes the representations from both image patches and text words as the input tokens.
$\mathbf{f}_{i,j}^p$ and $\mathbf{f}^q_{i,j}$ are the output feature representation vectors of the $j$-th token in the $i$-th layer with the clean input pair $(\mathbf{I}, \mathbf{T})$. $\hat{\mathbf{f}}_{i,j}^p$ and $\hat{\mathbf{f}}^q_{i,j}$ denote the output feature representation vectors of the $j$-th neuron in the $i$-th layer with the perturbed input pair $(\hat{\mathbf{I}}_{m-1}, \hat{\mathbf{T}}_{m-1})$.

Let $\hat{\mathbf{I}}_m^\prime$ denote the output by optimizing Eq.~\eqref{eq:feature_attack} with the clipping technique, which is further used to generate an enhanced image perturbation in the following section. This step is shown in Algorithm~\ref{alg:overview} lines 4-6.

\subsubsection{LLM-based Perturbation Enhancement}
In the context of transferable attacks, it is common for the pre-trained source model $\mathcal{F}$ to exhibit notable dissimilarities when compared to the victim target model $\mathcal{S}$. Consequently, relying solely on perturbing the latent representations using Eq.~\eqref{eq:feature_attack} may prove insufficient in ensuring the creation of high-quality adversarial samples capable of effectively attacking $\mathcal{S}$. To tackle this challenge, we present a solution that leverages the capabilities of Large Language Models (LLMs), such as ChatGPT~\cite{gpt4}, and the corresponding answers $\mathcal{Y}$ to bolster the process of perturbation generation.

$\bullet$ {\textbf{Masked Text Generation with LLM.}}
In a given visual-question pairing, multiple correct answers can exist, represented as $\mathcal{Y} = [\mathbf{y}_1, \cdots, \mathbf{y}_N]$, where $N$ corresponds to the count of correct answers. The primary objective of the transferable attack is to create adversarial instances in such a manner that the output of $\mathcal{S}(\hat{\mathbf{I}},\hat{\mathbf{T}})$ does not belong to the set $\mathcal{Y}$. To maximize the effectiveness of this transferable attack, a straightforward approach could involve compelling the pre-trained model $\mathcal{F}$ to produce incorrect predictions at each iteration. More specifically, this would entail ensuring that $\mathcal{F}(\hat{\mathbf{I}}_{m}^\prime,\hat{\mathbf{T}}_{m-1}) \notin \mathcal{Y}$.

However, it is important to note that this approach is impractical for the current state of pre-trained models $\mathcal{F}$, as they are not explicitly designed for predicting VQA answers during their pre-training phase. Fortunately, a viable alternative stems from the fact that many of these models incorporate the masked language modeling (MLM) task as part of their pre-training. In this context, we can transform the answer prediction task into a masked answer recovery task using the MLM framework.

Towards this end, we need to combine the perturbed question $\hat{\mathbf{T}}_{m-1}$ and each correct answer $\mathbf{y}_i \in \mathcal{Y}$ with a predefined prompt $\mathcal{P}$ using LLMs. Let $\hat{\mathbf{Z}}_{m, i} = \text{LLM}(\hat{\mathbf{T}}_{m-1}, \mathbf{y}_i, \mathcal{P})$ denote the combined sentence for the $i$-th correct answer. 
The next step is to mask the answer $\mathbf{y}_i$ from the generated sentence $\hat{\mathbf{Z}}_{m, i}$. Note that each answer $\mathbf{y}_i$ may contain multiple words. Let $\mathcal{M}_{i}$ denote the set of masked indices, and we can use $\hat{\mathbf{Z}}_{m, i\backslash{\mathcal{M}}_{i}}$ to represent the masked sentence.

$\bullet$ {\textbf{Masked Answer Anti-Recovery.}}
To achieve the transferable attack, we will prevent the model from recovering the correct answer tokens for each masked text $\hat{\mathbf{Z}}_{m, i\backslash{\mathcal{M}}_{i}}$, by minimizing the following anti-recovery loss:
\begin{equation}\label{eq:llm_loss}
    \mathcal{L}_{a}^m=\sum_{i=1}^{N}\sum_{j\in \mathcal{M}_i}\log(p_c(z_{m, i,j}|\hat{\mathbf{Z}}_{m,i\backslash{\mathcal{M}}_{i}},\hat{\mathbf{I}}_m^\prime)),
\end{equation} 
where $z_{m, i,j}$ is the $j$-th token in $\hat{\mathbf{Z}}_{m, i}$, and $p_c$ is the conditional probability score generated from the MLM head the pre-trained model $\mathcal{F}$ that is composed of a fully-connected layer and a softmax layer. 
After optimizing Eq.~\eqref{eq:llm_loss}, we clip the learned image perturbation again, and the output is denoted as $\hat{\mathbf{I}}_m^*$. This step is shown in Algorithm~\ref{alg:overview} lines 7-10.

\subsection{Cross-modal Joint Attack}
Due to the differences of input image $\hat{\mathbf{I}}_m^*$ and text $\hat{\mathbf{T}}_{m-1}$, we cannot use a unified approach to update their perturbations. For the numerical image, we can still use gradients and the clipping technique to update the perturbation, but for the discrete text, we propose to use the word substitution technique to replace words in the text with the help of continuous word embeddings.

\subsubsection{Joint Attack Trigger}
As discussed before, updating the text perturbation at each iteration is unnecessary. We design a heuristic function to determine when to trigger the joint attack by taking the number of informative words in the text (denoted as $|\mathcal{W}|$) and the maximum iterations $M$ into consideration. When $m \mod \lfloor\frac{M}{|\mathcal{W}|+1}\rfloor=0$, then {\name} triggers the joint attack. Here, the ``$+1$'' operation is to prevent attacking $\hat{\mathbf{T}}_{m-1}$ only in the last iteration step. The trigger is shown in Algorithm~\ref{alg:overview} line 12.
Otherwise, {\name} will output $(\hat{\mathbf{I}}_m^*, \hat{\mathbf{T}}_{m-1})$ as $(\hat{\mathbf{I}}_m, \hat{\mathbf{T}}_{m})$ for the $m$-the iteration (Algorithm~\ref{alg:overview} line 20).

Next, we introduce how to identify informative words and extract their synonyms.
Given a clean text ${\mathbf{T}}$, we first tokenize it and filter out all stop words using the Natural Language Toolkit (NLTK)\footnote{\url{https://www.nltk.org/}}, which results in a set $\{t_i|i\in\mathcal{W}\}$, where $\mathcal{W}$ represents the indices of the unfiltered tokens. 
For each token $t_i$, we follow BERT-Attack~\cite{bertattack} and employ the BERT model~\cite{bert} to predict the top-$K$ candidate words that share similar contexts, which results in a set of candidate words $\{c_{i,1},\cdots,c_{i,K}\}$. We then obtain the candidate set for all tokens $i\in\mathcal{W}$, and obtain a set $\mathcal{C}=\{c_{i,j}|1\leq j\leq K\}_{i\in\mathcal{W}}$. The motivation for using BERT is that it can better capture the context of a word, compared to other methods like Glove~\cite{glove} and Word2Vec~\cite{Word2Vec}. These candidates can retain more accurate syntactic and semantic information, making them more likely to satisfy semantic constraints. Note that this step can be done during ``Initialization'', which is fixed during word substitutions.

\subsubsection{Cross-modal Perturbation Generation}
After triggering the cross-modal attack, {\name} will update the gradients with regard to both perturbed image and text via minimizing the following latent feature-level loss function:
\begin{equation}\label{eq:cross_attack}
    \mathcal{L}_{c}^m = \mathcal{L}_{f}^m + \underbrace{\sum_{i=1}^{L_t} \sum_{j=1}^{D_t}Cos(\mathbf{f}_{i,j}^{t},\;\hat{\mathbf{f}}_{i,j}^{t})}_{\text{text encoder}},
\end{equation}
where $\mathcal{L}_f^m$ is the loss function from image and multimodal encoders with Eq.~\eqref{eq:feature_attack} using $(\hat{\mathbf{I}}_{m}^*, \hat{\mathbf{T}}_{m-1})$ and their corresponding token representations as the inputs, respectively. The second loss term is used to measure the feature similarity from the text encoder. 
$L_t$ denotes the mumbler of layers in the text encoder, and $D_t$ represents the number of input word tokens. 
$\mathbf{f}_{i,j}^t$ and $\hat{\mathbf{f}}_{i,j}^t$ denote the output feature representation vectors of from the clean input $(\mathbf{I}, \mathbf{T})$ and the perturbed input pair $(\hat{\mathbf{I}}_{m}^*, \hat{\mathbf{T}}_{m-1})$, respectively.

$\bullet$ {\textbf{Image Perturbation Update.}}
Since the image perturbations are numerical values, we can directly calculate the gradients using Eq.~\eqref{eq:cross_attack} and then apply the clipping technique to generate the output $\hat{\mathbf{I}}_m$ for the $m$-th iteration, as shown in Algorithm~\ref{alg:overview} lines 13-15.

$\bullet$ {\textbf{Text Perturbation Update.}}
Due to the discrete nature of text words, we need to unitize the learned gradients with Eq.~\eqref{eq:cross_attack} in the latent word embedding space to generate text perturbations motivated by~\cite{ye2022texthoaxer}. Toward this end, we propose to use word substitution attacks to generate text perturbations. 

\ul{\emph{Latent Word Embedding Estimation}.} 
The word substitution attack aims to replace the original, informative words in text $\hat{\mathbf{T}}_{m-1}$ with their synonyms, i.e., the words in set $\mathcal{C}$. To this end, we need to estimate the word representations after the attack first using the original informative word embeddings $\mathbf{E}(t_i)$ ($i \in \mathcal{W}$) and its gradient $\nabla\mathcal{L}_{c}^m(t_i)$ learned by Eq.~\eqref{eq:cross_attack} as follows:
\begin{equation}\label{eq:word_embedding}
    {\mathbf{E}}(\hat{t}_i) = \mathbf{E}(t_i) + \nabla\mathcal{L}_{c}^m(t_i).
\end{equation}

\ul{\emph{Synonym Ranking}.} 
The goal of synonym substitution is to find a synonym of $t_i$ from $\{c_{i,1}, \cdots, c_{i,K}\}$ to replace the original informative word $t_i$ and make the embedding of the synonym close to ${\mathbf{E}}(\hat{t}_i)$. 
Since there may be several informative words in $\mathcal{W}$, we need to decide the order of replacement. Intuitively, the larger similarity between ${\mathbf{E}}(\hat{t}_i)$ and the embedding of a synonym $c_{i,j}$, the higher chance of $c_{i,j}$ being a perturbation. To this end, we replace the original word with each synonym $c_{i,j}$ to generate each synonym's context-aware word embedding $\mathbf{E}(c_{i,j})$.
We then calculate the pair-wise cosine similarity between the estimated latent representation and the synonym context-aware word embedding as follows:
\begin{equation}\label{eq:ranking}
    \gamma_{i,j} = Cos({\mathbf{E}}(\hat{t}_i), \mathbf{E}(c_{i,j})).
\end{equation}

According to the similarity score values, we rank all the synonyms in $\mathcal{C}$ in descending order, denoted as $\mathcal{R}(\mathcal{C})$.

\ul{\emph{Synonym Substitution}.} 
We replace the original word in $\hat{\mathbf{T}}_{m-1}$ with its synonym that has the largest similarity in $\mathcal{R}(\mathcal{C})$. Let $\hat{\mathbf{T}}_{m-1}^\prime$ denote the new text sample. Then we check whether the new sample $\hat{\mathbf{T}}_{m-1}^\prime$ satisfies the constraint listed in Eq.~\eqref{eq:constraints}. 
If $Cos(U(\hat{\mathbf{T}}_{m-1}^\prime), U(\mathbf{T})) > \sigma_s$, then we keep the replacement in $\hat{\mathbf{T}}_{m-1}$, remove all the other synonyms of this word in $\mathcal{R}(\mathcal{C})$, and move to the next informative word.
If $Cos(U(\hat{\mathbf{T}}_{m-1}^\prime), U(\mathbf{T})) \leq \sigma_s$, we do not conduct the replacement and use the synonym with the second largest value in $\mathcal{R}(\mathcal{C})$. 

We will repeat this procedure until all informative words are replaced or all synonyms in $\mathcal{R}(\mathcal{C})$ are checked.
The output from this step is the perturbed text $\hat{\mathbf{T}}_m$ as shown in Algorithm~\ref{alg:overview} lines 16-19.

After executing all the above steps for $M$ iterations, we generate the final perturbed image-text pair $(\hat{\mathbf{I}}_M,\hat{\mathbf{T}}_M)$, which will be fed into different unknown victim models to conduct the transferable adversarial attack.

\section{Experiments}
\subsection{Experimental Setup}
\begin{table*}[t]
\setlength{\tabcolsep}{1.0mm}
    \centering
    \small
    \begin{tabular}{c|c|cc|ccc|cc||cc|ccc|cc}
    \toprule[1pt]
     \multirow{3}{*}{\makecell[c]{Source\\Model}} &\multirow{3}{*}{\makecell[c]{Target\\Model}} &\multicolumn{7}{c||}{VQAv2} & \multicolumn{7}{c}{TextVQA}\\\cline{3-16}
     && \multicolumn{2}{c|}{Text Only} & \multicolumn{3}{c|}{Image Only} & \multicolumn{2}{c||}{Multi-modality}& \multicolumn{2}{c|}{Text Only} & \multicolumn{3}{c|}{Image Only} & \multicolumn{2}{c}{Multi-modality} \\\cline{3-16}
     & & B\&A & R\&R & DR & FDA & SSP & CoA & {\name} & B\&A & R\&R & DR & FDA & SSP & CoA & {\name} \\\hline

     \multirow{5}{*}{{ViLT}}
     &ALBEF &10.28&5.20&8.78&9.84&24.90&16.70&\textbf{30.36}&13.00&5.80&8.20&9.40&17.00&15.40&\textbf{22.20}\\
     &TCL &11.86&6.08&8.74&9.62&22.54&17.84&\textbf{27.96}&12.20&4.80&7.10&8.20&13.60&14.90&\textbf{19.80}\\
     &VLMO-B &6.34&1.82&5.08&5.70&21.48&13.64&\textbf{25.72}&7.30&3.20&7.40&5.70&13.90&12.90&\textbf{19.50}\\
     &VLMO-L &5.02&2.18&5.58&5.72&13.08&10.64&\textbf{25.98}&6.60&0.30&2.80&2.40&7.20&7.60&\textbf{8.40} \\ \hline\hline
     
    \multirow{5}{*}{{TCL}} 
     & ViLT & 6.68 & 2.52 & 5.74 & 5.78 & 11.04 & 11.22 & \textbf{21.80} & 9.30 & 2.60 & 4.60 & 5.20 & 7.10 & 10.80 & \textbf{16.30} \\
     & ALBEF & 5.58 & 2.92 & 11.10 & 12.52 & 38.26 & 33.24 & \textbf{58.42} & 10.80 & 8.70 & 9.10 & 10.50 & 31.80 & 26.10 & \textbf{46.80} \\ 
     & VLMO-B & 7.52 & 3.84 & 15.82 & 9.00 & 23.88 & 18.32 & \textbf{47.48} & 7.82 & 2.54 & 6.50 & 7.60 & 16.70 & 15.50 & \textbf{34.00} \\ 
     & VLMO-L & 5.64 & 2.22 & 8.04 & 6.14 & 15.26 & 12.64 & \textbf{30.46} & 2.40 & 5.96 & 3.80 & 4.70 & 9.50 & 10.00 & \textbf{18.60} \\ 
     \hline\hline

     \multirow{5}{*}{{ALBEF}} 
     & ViLT &6.72 &2.42 & 6.90& 7.02& 11.42& 11.36 & \textbf{21.60} &8.70 &2.60 & 4.60& 5.80& 8.20& 11.70 & \textbf{15.60} \\ 
     & TCL & 6.96 & 1.80 & 12.64 & 11.78 & 35.46 & 27.24 & \textbf{61.32} & 9.90 & 2.90 & 9.60 & 8.80 & 13.10 & 20.50 & \textbf{43.70} \\
     & VLMO-B & 5.68 & 2.04 & 8.14 & 9.04 & 21.48 & 16.16 & \textbf{42.32} & 8.50 & 3.30 & 7.70 & 8.10 & 15.20 & 14.50 & \textbf{28.30} \\
     & VLMO-L & 5.02 & 2.18 & 5.58 & 5.72 & 21.56 & 10.64 & \textbf{25.98} & 5.70 & 2.20 & 4.10 & 4.50 & 8.20 & 7.40 & \textbf{16.20} \\ 
     \hline\hline

     \multirow{4}{*}{VLMO-B}
     &ViLT &7.72&2.04&4.36&5.34&10.20&10.90&\textbf{18.70}&10.90&0.80&3.20&3.40&7.80&11.70&\textbf{15.20}\\
     &TCL &12.20 &6.26&10.98&13.64&20.24&21.52&\textbf{43.62}&13.50&4.50&8.20&9.30&14.30&18.00&\textbf{28.30}\\
     &ALBEF &10.74&6.30&11.22&14.52&22.66&22.46&\textbf{48.06}&13.50&6.10&9.50&12.70&16.80&19.60&\textbf{32.60}\\
     &VLMO-L&5.98&3.96&4.58&5.48&10.66&12.52&\textbf{30.82}&6.70&0.60&2.70&4.20&6.80&9.60&\textbf{17.40} \\ 
     \hline\hline

     \multirow{4}{*}{VLMO-L}
     &ViLT &7.50&1.62&7.48&3.52&7.94&8.78&\textbf{13.08}&10.30&1.30&3.00&2.90&5.80&9.20&\textbf{13.10}\\
     &TCL &12.20&6.14&12.10&10.92&21.18&15.48&\textbf{32.96}&12.90&4.40&6.90&6.80&15.60&13.70&\textbf{21.70}\\
     &ALBEF &10.84&5.98&24.84&10.90&24.50&15.14&\textbf{37.48}&13.00&6.40&9.30&9.40&17.00&12.30&\textbf{26.80}\\
     &VLMO-B &8.22&1.86&20.96&7.58&19.60&12.70&\textbf{33.78}&8.70&1.90&6.00&4.50&14.20&11.60&\textbf{25.20} \\ 
     \hline

    \end{tabular}
    \caption{Comparison between {\name} and baselines on different models using the VQAv2 and TextVQA datasets (\%).}
    \label{tab:main_results}
\end{table*}
\noindent\textbf{\ul{Datasets $\&$ Models}} We evaluate the proposed {\name} on the VQAv2~\cite{vqav2} and TextVQA~\cite{textvqa} datasets. We randomly select \textbf{6,000} and \textbf{1,000 correctly predicted samples} from the VQAv2 and TextVQA validation datasets, respectively. Because an image-question pair may have multiple candidate answers provided by crowd workers, we define a correct prediction only if the predicted result is the same as the label with the highest VQA
score\footnote{VQA score calculates the percentage of the predicted answer that appears in 10 reference ground truth answers. More details can be found via  \url{https://visualqa.org/evaluation.html}}. Each selected sample is correctly classified by all target models.  We also development experiments on \textbf{five models}, including ViLT~\cite{vilt}, TCL~\cite{tcl}, ALBEF~\cite{albef}, VLMO-Base (VLMO-B)~\cite{vlmo}, and VLMO-Large (VLMO-L)~\cite{vlmo}. Note that VLMO-B and VLMO-L share the same structure but have different model sizes. These models are first pre-trained on public image-text pairs and then fine-tuned on VQA datasets.

\noindent\textbf{\ul{Baselines} }
We comprehensively compare {\name} with text, image, and multi-modal adversarial attack methods.
Specifically, we first adopt BERT-Attack (B\&A)~\cite{bertattack} and Rewrite-Rollback (R$\&$R)~\cite{randr} as \textbf{text-attack} baselines. 
For \textbf{image attack} methods, we adopt DR~\cite{DRattack}, SSP~\cite{SSP}, and FDA~\cite{FDA} as baselines. These methods generate adversarial images by only perturbing intermediate features and can thus be directly utilized in our problem. 
{\name} is also compared with the\textbf{ multi-modal attack} approach Co-Attack (CoA)~\cite{coattack}. To the best of our knowledge, it is the only scheme that attempts to simultaneously add image and text perturbations.  
We adopt the \emph{Attack Success Rate} (\textbf{ASR}) as the evaluation metric, which measures the ratio of samples whose predicted labels are not in the correct answers. 

\subsection{Result Analysis}

\begin{figure}[t]
\begin{center}
\includegraphics[width=0.90\linewidth]{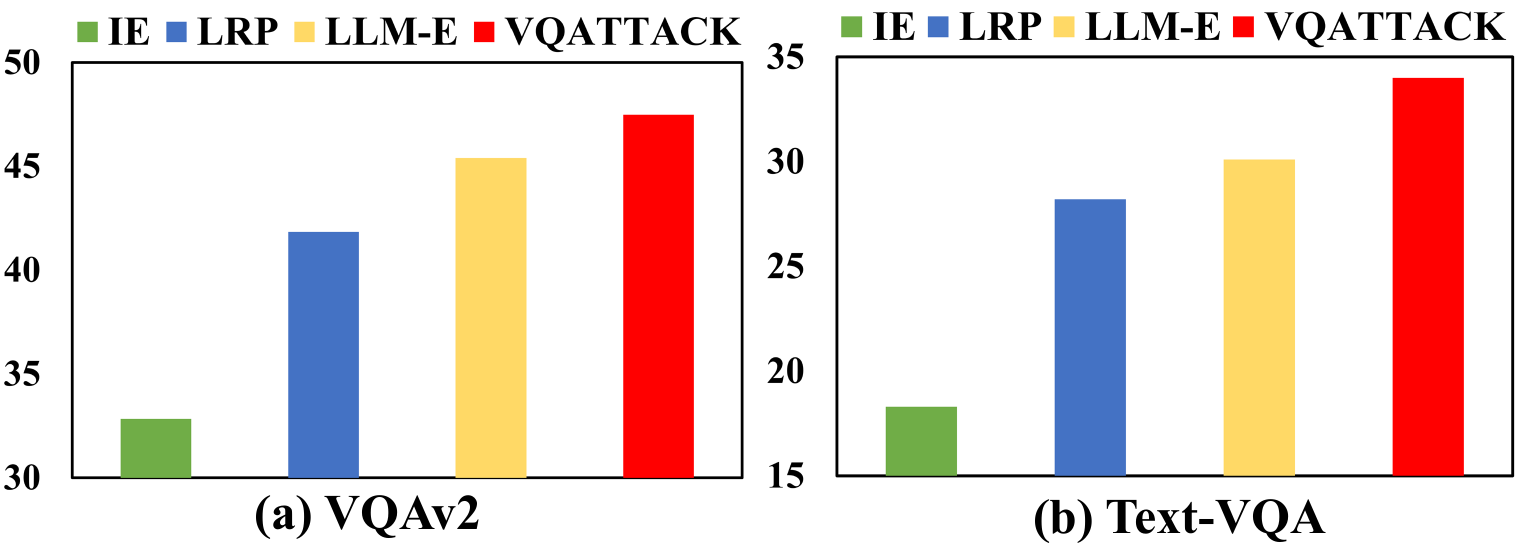}
\end{center}
\caption{Ablation study results on the source model TCL and the victim model VLMO-B.}
\label{fig:ab}
\end{figure}
We alternatively select a pre-trained model as the source model to generate adversarial samples, which are then used to attack the remaining models treated as victims.
Experimental results are listed in Table~\ref{tab:main_results}.
We can observe that the proposed {\name} significantly outperforms all baselines on each dataset for the five transferable attack experiments. 
Specifically, {\name} achieves an average ASR of 22.49$\%$ using ViLT as the source model, 34.23$\%$ for TCL, 31.88$\%$ for ALBEF, 29.33$\%$ for VLMO-B and 25.51$\%$ for VLMO-L. 
The ASR value is comparatively lower when using ViLT as the source model because its model structure and pre-training strategies are greatly different from others. Also, the ASR value obtained by using VLMO-L as the source model is slightly lower than that of using VLMO-B as the source model. This observation demonstrates that the model owns larger parameters can present better adversarial robustness.
Finally, all of these results have demonstrated the effectiveness of our proposed approach and also comprehensively reveal the huge threat of adversarial attacks in the ``pre-training $\&$ fine-tuning'' learning paradigm.




\subsection{Ablation Study}
This ablation study aims to validate the effectiveness of the two designed modules. 
Figure~\ref{fig:ab} shows the ablation study results using the adversarial samples generated by TCL to attack VLMO-B. ``IE'' means only using the latent presentations learned by the image encoder to generate adversarial samples in Eq.~\eqref{eq:feature_attack}. ``LRP'' means the latent representation perturbation used in the LLM-enhanced image attack module, where we only use Eq.~\eqref{eq:feature_attack} to generate the adversarial samples. We can observe that using the multimodal encoder can make significant ASR improvements. ``LLM-E'' means using both Eqs.~\eqref{eq:feature_attack} and~\eqref{eq:llm_loss} to generate perturbations. Compared with ``LRP'', the performance can increase, which indicates the efficacy of introducing LLM to help generate masked text and the effectiveness of the designed masked answer anti-recovery loss in Eq.~\eqref{eq:llm_loss}. The proposed {\name} achieves the best performance. The performance gap between LLM-E and {\name} demonstrates the effectiveness of the proposed cross-modal joint attack module.


\begin{figure}[t]
\begin{center}
\includegraphics[width=0.90\linewidth]{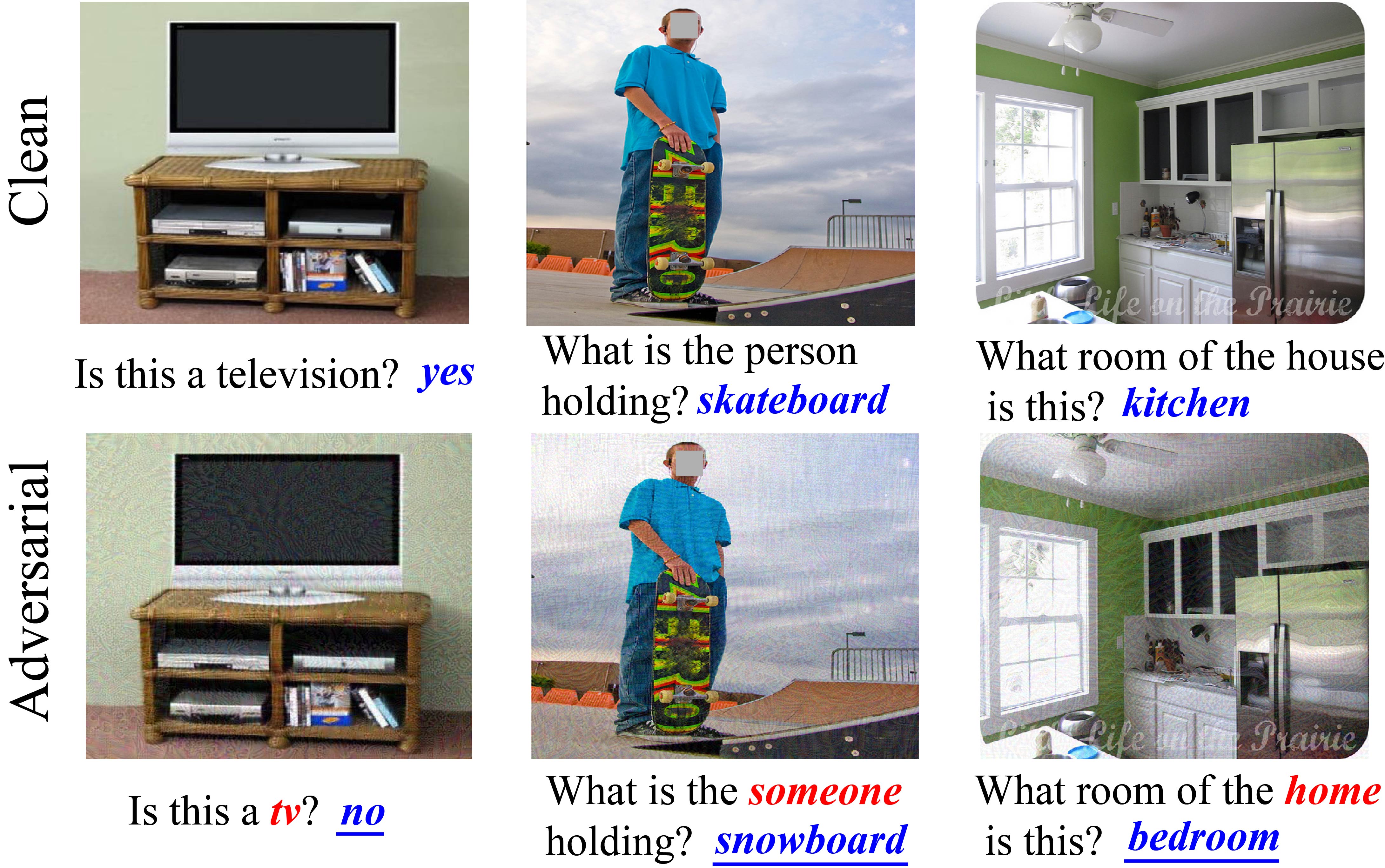}
\end{center}
\caption{Qualitative results of {\name} on the VQAv2 dataset generated by the TCL model. The original answer and perturbed words are displayed in blue and red, respectively. The wrong prediction is shown with an underline.}
\label{fig:case}
\end{figure}

\begin{table*}[t]
\setlength{\tabcolsep}{1.30mm}
    \centering
    \small
    \begin{tabular}{c|c|ccccc|ccccc}
    \toprule[1pt]
     \multirow{2}{*}{\makecell[c]{Modality}} &\multirow{2}{*}{\makecell[c]{Method}} &\multicolumn{5}{c|}{VQAv2} & \multicolumn{5}{c}{TextVQA} \\\cline{3-12}
    & &ViLT& ALBEF & TCL &  VLMO-B& VLMO-L & ViLT&ALBEF & TCL &  VLMO-B& VLMO-L \\ \hline 
     \multirow{2}{*}{\makecell[c]{Text \\ Only}}   
     & B\&A & 15.16 & 8.24   &8.96 &10.16&11.72&20.20&11.50&14.90&13.00&10.10  \\
      & R\&R & 7.30 & 4.68   &5.64&6.86&4.62&7.40&8.30& 5.90&2.70&3.80  \\\hline
      \multirow{3}{*}{\makecell[c]{Image \\ Only}}
     &DR&  16.90 &  20.42  &15.82&17.12&11.02&14.40&14.50&11.60&14.50&7.90\\
      &FDA& 20.08 & 17.72   &16.74&22.16&9.92&13.90&12.80&11.70&19.50&7.50\\
      &SSP& 61.36 & 49.68  &51.46&46.32&41.94&49.80&36.70&40.70&34.60&28.40\\\hline
     \multirow{2}{*}{\makecell[c]{Multi-\\modality}} 
     &Co-Attack&50.12&46.50&52.74&43.56&18.48&42.30&35.80&45.80&35.80&16.80\\
     &{\name}&\textbf{79.00}&\textbf{75.16}&\textbf{76.46}&\textbf{75.04}&\textbf{61.60}&\textbf{65.00}&\textbf{61.90}&\textbf{65.70}&\textbf{66.20}&\textbf{48.70}\\\hline
     
    \end{tabular}
    \caption{Results of transferable attacks between $\mathcal{F}$ and $\mathcal{S}$ with the same pre-trained structures.}
    \label{tab:main_results_TAS}
\end{table*}




\subsection{Case Study}
We conduct a case study on the VQAv2 dataset using the source model TCL, as shown in Figure~\ref{fig:case}. We can observe that the generated adversarial samples largely change the original correct prediction to a wrong answer. For instance, recognizing a kitchen as a bedroom (column 3). Furthermore, the generated adversarial samples still keep the natural appearance as the benign samples, which demonstrates a serious security threat in the present VQA systems.

\subsection{Transferable Attacks with Shared Information}
 In this experiment, we use the pre-trained model $\mathcal{F}$ as the source model and its downstream VQA task as the target $\mathcal{S}$. $\mathcal{F}$ and $\mathcal{S}$ share most of the structures, and only the final prediction layers are different. 
Table~\ref{tab:main_results_TAS} shows the experimental results. We can observe that the proposed {\name} still outperforms all the baselines on the two VQA datasets. Compared with the results listed in Table~\ref{tab:main_results}, we can observe that the performance of all approaches improves significantly under this setting. This experiment concludes that the shared information is sensitive, which may make the target models vulnerable.
\section{Related Work}

\textbf{Robustness of VQA} 
The robustness of VQA is moderately explored. 
Recently, Fool-VQA~\cite{foolvqa} explores the adversarial vulnerability of a VQA system by adding image noise constrained by $l_{\infty}$ distance.
TrojVQA~\cite{trojvqa} performs a backdoor
attack by injecting deliberate image patches and word tokens. These studies concentrate on the robustness of end-to-end trained VQA models and design algorithms
based on the final predictions. Because the outputs of pre-trained and fine-tuned VQA models are different, they cannot be extended to our problem.

\noindent\textbf{Adversarial Attacks}
Adversarial image attacks are
initially explored in Fast Gradient Sign Method~\cite{fgsm} and Projected Gradient Decent~\cite{pgd}. 
An intriguing property of these adversarial images is their ``transferability'', which can be utilized to attack different image models with unknown parameters and structures. 
 To enhance the transferability, the recently proposed methods exploit features from intermediate layers for adversarial attacks. They either combine the feature distortion loss with the classification cross-entropy term~\cite{ILA,Fda2,fdafd} or fully rely on the intermediate feature disruption~\cite{FDA,SSP}. 
Text attack methods are primarily divided into searching-based and gradient-based algorithms. 
Searching-based attacks include a set of heuristic ranking algorithms~\cite{bertattack,randr,CLARE} with sub-optimal performance. Recently, gradient-based attacking approaches~\cite{GBDA,ye2022texthoaxer,SEMattack,ye2022leapattack} are proposed. Unlike image attacks, the gradient cannot be directly projected onto {discrete} text inputs. Accordingly, gradient change is instantiated either through distance matching on candidate word embeddings~\cite{ye2022texthoaxer, ye2022leapattack,SEMattack}, or by using Gumbel-softmax sampling~\cite{gumbel} on a learnable distribution of all candidate words. 
For multi-modal attacks, the recently proposed Co-attack~\cite{coattack} method firstly combines both image and text attacks, which utilizes word substitution to guide image adversarial attacks. 
It has demonstrated to some extent that perturbations across both modalities can be more effective than a single source. However, it does not take into account the dynamic connections between perturbations on different modalities, indicating potential space for significant improvements under more challenging scenarios.

\section{Conclusion}
In this paper, we investigate a novel transferable adversarial attack scenario, aiming to generate adversarial samples only using pre-trained models, which are used to attack different black-box victim models. Correspondingly, we propose a new model named {\name}, which can jointly update both image and text perturbations. Besides, we propose to incorporate the large language model to enhance the transferability of the source model. Experimental results on two VQA datasets with five models show the effectiveness of the proposed {\name} for the transferable attacks.

\section{Acknowledgements} This work is partially supported by the National Science Foundation under Grant No. 1951729, 1953813, 2119331, 2212323, and 2238275.

\bibliography{aaai24}
\clearpage
\section{Appendix}

\subsection{A. Details of LLM Utilization}

\begin{figure}[h]
         \centering
         \includegraphics[width=0.48\textwidth]{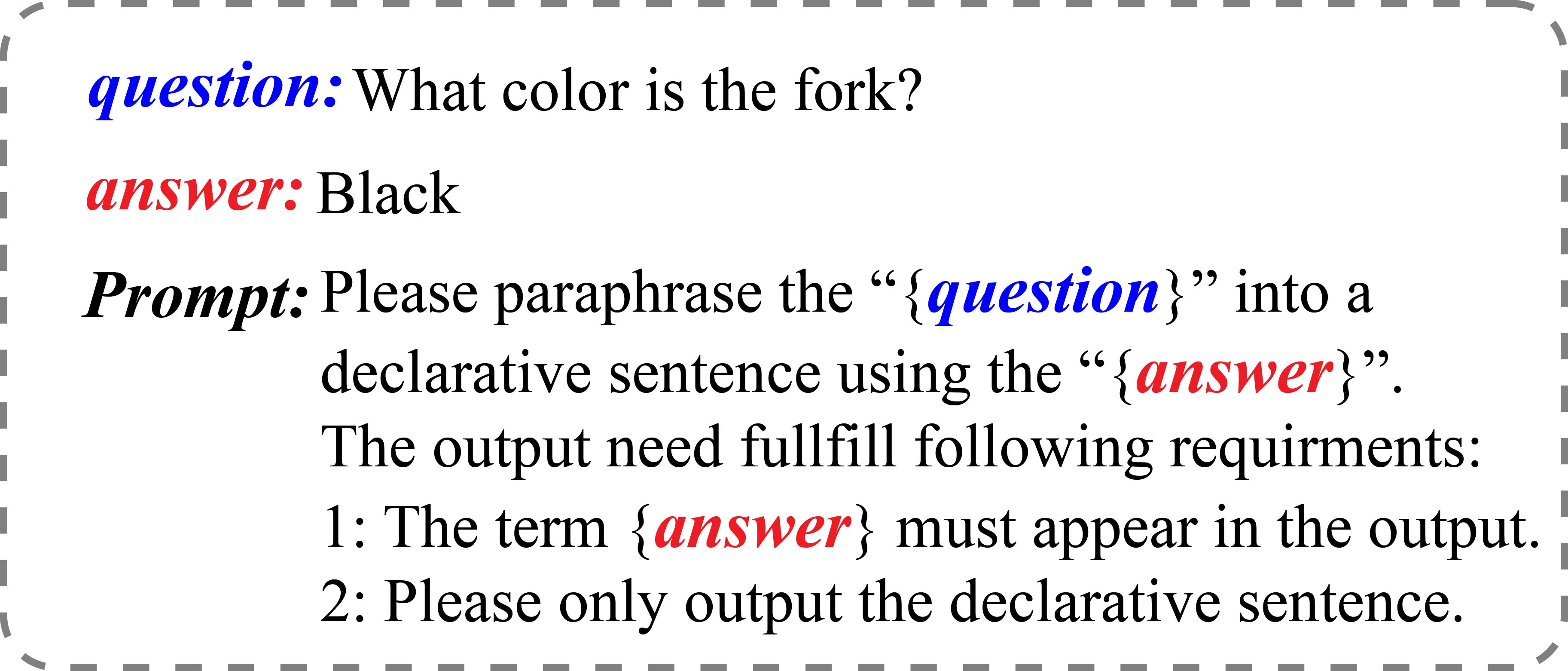}
         \caption{An example of Transferable adversarial attacks on VQA via pre-trained models.}
         \label{fig:LLM}
\end{figure}

In this section, we introduce how to use the LLM to combine the perturbed question $\hat{\mathbf{T}}_{m-1}$ and a correct answer $\mathbf{y}_i$ into a sentence. We use ChatGPTv4~\cite{liu2023summary} to accomplish this target because it has demonstrated stronger language reasoning capabilities compared to other LLMs. 
When utilizing ChatGPT-v4, we first design a prompt $\mathcal{P}$ as illustrated in Figure~\ref{fig:LLM}.

As illustrated in Figure 1, the prompt consists of three parts. The first is the task description, in which we need to combine ``\{\textcolor{blue}{question}\}'' and ``\{\textcolor{red}{answer}\}'' into a declarative sentence. To ensure the quality of the combined sentences, we add the following two constraints:
\begin{itemize}
    \item \underline{\emph{The term \{\textcolor{red}{answer}\} must appear in the output}}. This constraint emphasizes that the output sentence must contain the answer, which is a prerequisite for masked text generation.
    \item \underline{\emph{Please only output the declarative sentence}}. This constraint aims to avoid redundant prompts and prefixes, such as ``\emph{Sure, here is the output sentence...}''. 
Such a prefix is unnecessary and may even introduce interference in the masked answer anti-recovery step, thus affecting the attack performance.
\end{itemize}

\subsection{B. Details of VL Models}
In this section, we illustrate the details of all five VL models evaluated in our paper, including ViLT~\cite{vilt}, ALBEF~\cite{albef}, TCL~\cite{tcl}, VLMO-B and VLMO-L~\cite{vlmo}. These models consist of an image encoder, a text encoder, and a multimodal encoder.
\begin{figure*}[t]
         \centering
         \includegraphics[width=0.95\textwidth]{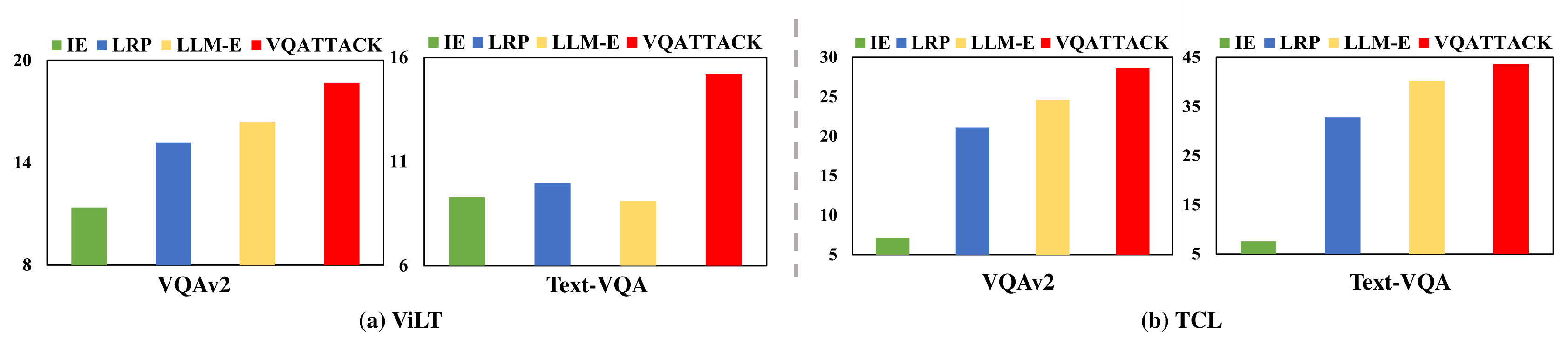}
         \caption{More examples of ablation study. We generate adversarial samples form the pre-trained VLMO-B model, and use the output to attack ViLT and TCL.}
         \label{fig:appendix}
\end{figure*}
\begin{itemize}[leftmargin=*]
    \item \textbf{ViLT}. We select the model ViLT because it employs a succinct model structure and significantly outperforms previous end-to-end trained VQA models on the VQAv2 dataset. The image encoder of ViLT is a linear projection layer. Given an input image $\mathbf{I}$, it first divides the image into patches with equal spatial resolution. Then each image patch is flattened into a vector and encoded by a linear transformation, which results in $D_{p}$ image tokens. For the text $\mathbf{T}$, it is first tokenized and embedded by the commonly used byte-pair encoder. The word embeddings are then encoded through a text encoder, which is a word-vector linear projection layer. The latent token representations from image and text encoders are then concatenated with a learnable special token $\langle cls\rangle$ and fed into the multi-modal encoder, which is a twelve-layer transformer encoder. The encoder attends tokens of different modalities through the self-attention mechanism. At the pre-training
    stage, the output features of text tokens are fed into the
    MLM head to predict the masked word tokens. 
    When fine-tuning on VQA task, the output feature from the $\langle cls\rangle$ token is fed into a VQA prediction head to select the correct answer. The VQA prediction head is composed of a fully-connect layer and a softmax layer.
 
    \item \textbf{ALBEF.} The ALBEF model has a different structure from ViLT. Specifically, the image encoder is a twelve-layer visual transformer ViT-B/16~\cite{vit}. The text encoder is a six-layer transformer encoder. The structure of the multi-modal encoder is the same as a six-layer transformer decoder, where each layer contains a self-attention module, a cross-attention module and a feed-forward module. After obtaining the image and text token features through the image/text encoders, the multi-modal encoder first accepts the text token features as input and attends to them through the self-attention module. Then, the output will fuse with image token features through the cross-attention module. At the pre-training stage, the output features from the multi-modal encoder will be processed by the MLM head.  For prediction on the VQA task, the multi-modal features are fed into a six-layer transformer decoder. The decoder also accepts a sequence initialized by the $\langle cls\rangle$ token as input and interacts with the multi-modal representations through cross-attention layers. As a result, the VQA answer is auto-regressively generated in an open-ended manner. 

    \item \textbf{TCL.} TCL follows the same structures as ALBEF but has more different pre-training tasks. In addition to the traditional pre-training tasks like MLM, it introduces additional tasks through contrastive learning. These contrastive learning tasks are developed from the uni-modal and cross-modal levels based on additional image-text pairs. The experimental results indicate that TCL improves the quality of multimodal representations by pre-training with these extra tasks. After fine-tuning TCL on the VQA task, its performance also notably outperforms that of ALBEF.

    \item \textbf{VLMO-B.} VLMO-B adopts a transformer structure for each encoder. Both the image and text encoders are one-layer transformer encoders, and the multimodal encoder is a twelve-layer transformer encoder.
    VLMO-B also adopts a modular design for each encoder, replacing the original feed-forward (FFN) layer with a modality-aware FFN head. At the pre-training stage, each encoder has different parameters in the modality-aware FFN head, while the multi-head self-attention module is shared between the image and text encoders.
    For the VQA prediction, the output feature from the  $\langle cls\rangle$ token is fed into a VQA prediction head to select the correct answer, which is the same as the ViLT model.  

    \item \textbf{VLMO-L.} We also adopt a larger version of VLMO-B to evaluate the transferable attack performance under different model sizes. VLMO-L has the same structure as VLMO-B but with more layers and parameters. Specifically, the multi-modal encoder is a transformer encoder with 24 layers, and each latent representation owns a size of 1,024 dimensions, which is 768 in VLMO-B. As a result, the parameters of VLMO-L are three times larger than those of VLMO-B (562M vs 175M).

\end{itemize}

\subsection{C. Implementation Details}
In this section, we show the implementation details of our experiments.
For the perturbation parameters on images, we follow the previous transferable image attack methods~\cite{ILA,tapattack} and set the $L_\infty$-norm distance threshold $\sigma_{i}$ to 16/255. 
Following the PGD, the maximum iteration steps $M$ is set to 20 on {\name} and 40 on all image attack baselines, including SSP, FDA, DR, and the multi-modal attack Co-Attack (CoA). 
This is because, in most steps, {\name} needs to update gradients twice in one iteration step, while other baselines do it only once. Thus, we set a smaller $M$ to our approach.
Finally, all methods compared in experiments are optimized with the DIM~\cite{diattack}, which is an image augmentation strategy and has been widely adopted in current transferable image attacks~\cite{DRattack,wu2020skip}.

For the text modality, we set the semantic similarity constraint $\sigma_s$ to 0.95,  which follows the previous text-attack work~\cite{randr,bertattack}. Because the text-attack baselines B$\&$A and R$\&$R need to do queries to the target model, which is different from our transferable attack setting. Thus,  when running B$\&$A and R$\&$R, we first generate adversarial texts by querying the VQA model fine-tuned on the source pre-trained one and then perform a transferable attack by sending the querying results to the victim model. 
Finally, all experiments are conducted on a single GTX A100 GPU.

\subsection{D. More Ablation Study Analysis}
As shown in Fig.~\ref{fig:appendix}, we also display more results of the ablation study. The adversarial examples are generated from the pre-trained ViLT source model, and they are transferred to attack the victim ViLT and TCL models. From the figure, we can observe that the performance is consistent with the analysis of the ablation study, except the ASR value of LLM-E is slightly lower than LRP on the ViLT model through the Text-VQA dataset. We attribute this to the huge differences in pre-training strategies on MLM tasks between VLMO-B and ViLT. Specifically, the ViLT model directly uses the MLM task to pre-trained the whole model from initialization. However, VLMO-B adopts a stage-wise pre-training strategy, which first pre-trains the image and text encoder on uni-modal tasks and then trains the multi-modal encoder on the MLM task with a good initialization of uni-modal representations. 
Finally, by combining the LLM-E image attack module with the cross-modal joint attack module, the performance of {\name} significantly surpasses that of individual components across all figures. This further demonstrates the effectiveness of our proposed method.

\end{document}